\documentclass[10pt,twocolumn,letterpaper]{article}

\usepackage{iccv}
\usepackage{times}
\usepackage{epsfig}
\usepackage{graphicx}
\usepackage{amsmath}
\usepackage{amssymb}
\usepackage{blindtext}
\RequirePackage{caption}
\usepackage{booktabs}
\usepackage{tabularx}
\usepackage{overpic}
\usepackage{wrapfig}
\usepackage{multirow}
\newcolumntype{Y}{>{\centering\arraybackslash}X}
\usepackage{xcolor}

\usepackage{pifont}
\usepackage[bottom]{footmisc}
\usepackage{algorithm}
\usepackage{algpseudocode}
\usepackage[accsupp]{axessibility}  
\usepackage{stfloats}

\let \bs=\boldsymbol

\def \vq{\bs{q}}

\def \saliency {\textup{\saliency}}

\def \path {\mathit{path}}

\usepackage{babel}
\usepackage{arydshln}
\date{}

\makeatother

\numberwithin{theorem}{section}
\numberwithin{lem}{section}

\usepackage[pagebackref=true,breaklinks=true,letterpaper=true,colorlinks,bookmarks=false]{hyperref}

\iccvfinalcopy 

\def\iccvPaperID{10278} 

\ificcvfinal\fi

\begin{document}

\title{LiDAR-Based 3D Object Detection via Hybrid 2D Semantic Scene Generation}

\author{Haitao Yang\textsuperscript{1} \hspace{0.2in}
Zaiwei Zhang\textsuperscript{2} \hspace{0.2in}
Xiangru Huang\textsuperscript{3} \hspace{0.2in}
Min Bai\textsuperscript{2} \hspace{0.2in}
Chen Song\textsuperscript{1}
\\
Bo Sun\textsuperscript{1} \hspace{0.3in}
Li Erran Li\textsuperscript{2} \hspace{0.3in}
Qixing Huang\textsuperscript{1}
\vspace{4pt}
\\
\textsuperscript{1}{The University of Texas at Austin}\hspace{0.3in} \textsuperscript{2}{AWS AI}\hspace{0.3in} \textsuperscript{3}{MIT CSAIL}
}

\twocolumn[{%
\renewcommand\twocolumn[1][]{#1}%
\maketitle
\begin{center}
\centering
\begin{tabular}{cc}
\includegraphics[width=1\columnwidth]{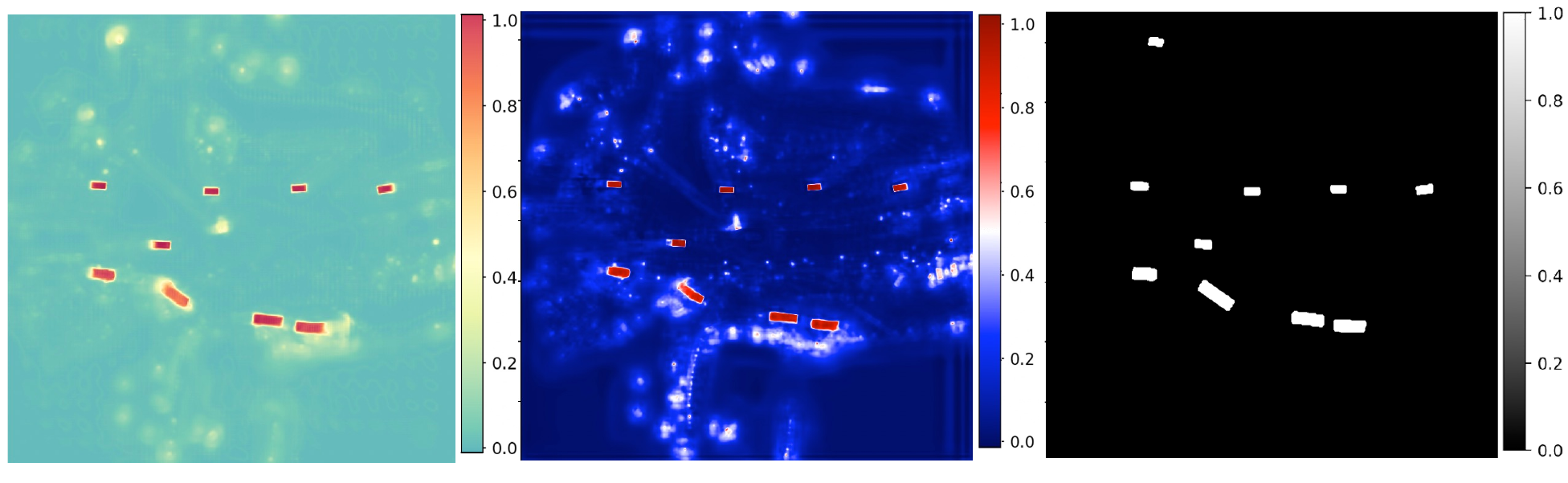} &
\includegraphics[width=1\columnwidth]{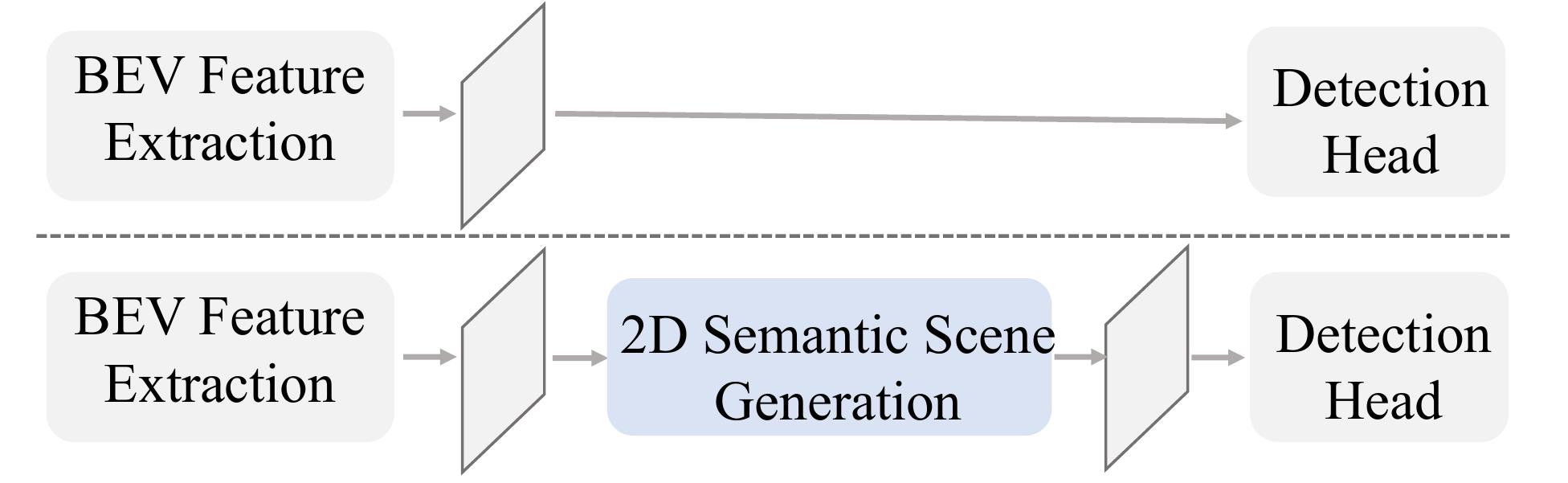}  \\
(a) 2D Semantic Scene & (b) Pipeline Comparison
\end{tabular}
\captionsetup{type=figure}
\captionof{figure}{(a) We propose to integrate the 2D semantic scene representation into LiDAR-based 3D object detectors, which can be generated by either an explicit network (left) or an implicit network (middle) using the projection of 3D bounding boxes into the Bird's Eye View (right) as supervision. (b) The proposed 2D semantic scene representation can be easily integrated into most existing detectors (top) as a 2D semantic scene generation module (bottom) with minimal engineering efforts.}
\label{Figure:Teaser}
\end{center}%
}]

\ificcvfinal\thispagestyle{empty}\fi

\begin{abstract}

Bird's-Eye View (BEV) features are popular intermediate scene representations shared by the 3D backbone and the detector head in LiDAR-based object detectors. However, little research has been done to investigate how to incorporate additional supervision on the BEV features to improve proposal generation in the detector head, while still balancing the number of powerful 3D layers and efficient 2D network operations. This paper proposes a novel scene representation that encodes both the semantics and geometry of the 3D environment in 2D, which serves as a dense supervision signal for better BEV feature learning. The key idea is to use auxiliary networks to predict a combination of explicit and implicit semantic probabilities by exploiting their complementary properties. Extensive experiments show that our simple yet effective design can be easily integrated into most state-of-the-art 3D object detectors and consistently improves upon baseline models.

\end{abstract}
\section{Introduction}
LiDAR-based object detection is a critical task in 3D recognition with important applications in autonomous driving and robotics. Unlike the depth sensor commonly used indoors, a LiDAR device is often mounted on a moving vehicle and is subject to dynamic outdoor conditions. The differences in sampling densities, object sizes, and scene scales pose a set of unique challenges, stimulating a wave of research efforts to design 3D object detection models specifically for LiDAR point clouds. 

Existing works in LiDAR-based 3D object detection typically use points~\cite{Shi_2019_pointrcnn, Yang_2020_3dssd, Zhang_2022_napae, Qi_2018_frustum, Qi_2019_votenet, Shi_2020_pointgnn, Yang_2019_std}, voxel grids~\cite{Zhou_2018_voxelnet, Yan_2018_second, Lang_2019_pointpillar, Deng_2021_voxelrcnn, chen_2022_focal}, or their hybrids~\cite{Shi_2020_pvrcnn, wang_2020_pillarod, Mao_2021_pyramidrcnn, Noh_2021_hvpr, He_2020_sassd, chen_2017_mv3d} as the input representation of the 3D scene. Point-based methods utilize PointNet and its variants~\cite{Qi_2017_pointnet, Qi_2017_pointnet2} as the 3D backbone to extract point-wise features directly from the input. In contrast, grid-based methods voxelize the raw point cloud by either simple discretization or point-based feature aggregation within each cell~\cite{Zhou_2018_voxelnet, Lang_2019_pointpillar}. Thanks to the sparse convolutional operation, grid-based methods are computationally efficient and usually outperform their point-based counterparts. At the time of paper writing, grid-based methods have a dominant role in LiDAR-based 3D object detection. 

\begin{figure}
\centering
\includegraphics[width=0.45\textwidth]{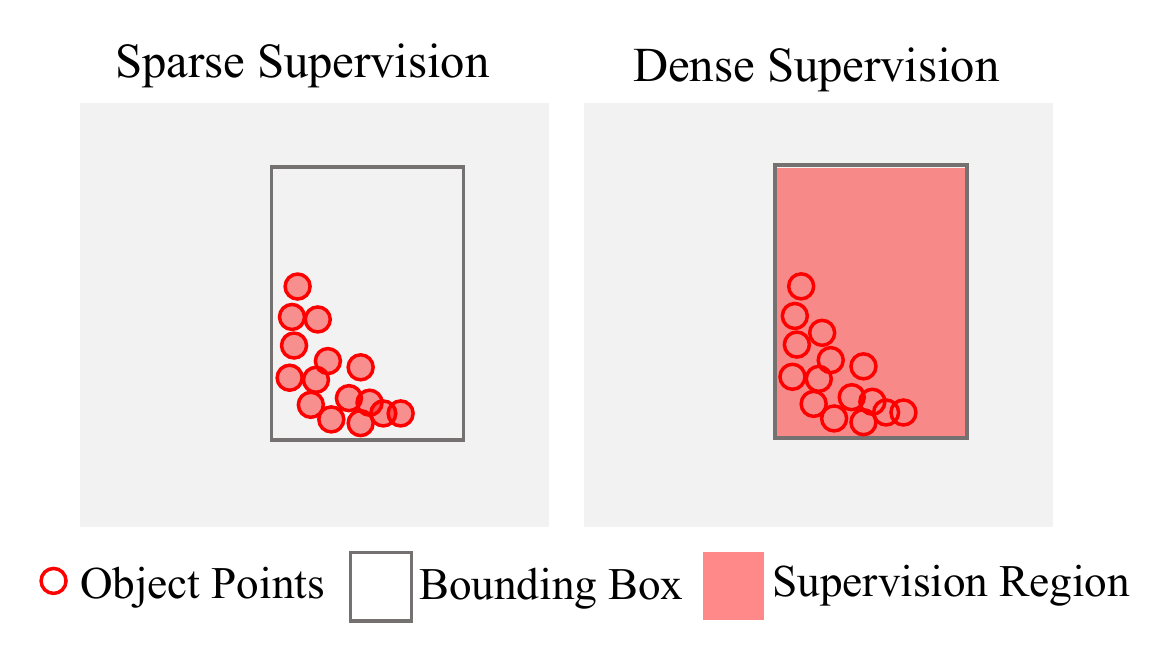}
\caption{When the ground-truth bounding box contains very few object points, the sparse supervision paradigm used by existing works (left) fails to provide strong enough regularization for object detection. In contrast, the proposed 2D semantic scene representation utilizes dense training supervision (right), covering interior locations without point samples.}
\label{Figure:Intro:dense-sup}
\end{figure}

This paper seeks to design an efficient and powerful LiDAR-based 3D object detector. A typical LiDAR-based 3D object detector first uses a 3D backbone to extract voxel features via sparse convolution~\cite{Zhou_2018_voxelnet} or pillar features via PointNet~\cite{Lang_2019_pointpillar} from the input. The extracted features are then projected to 2D, forming Bird's-Eye View (BEV) features. Finally, the detection head takes the BEV features as input and generates object proposals. In this pipeline, the BEV features serve as condensed representations of the 3D scene, designed to balance the number of powerful 3D layers and efficient 2D network operations. The quality of BEV features directly affects proposal generation and dictates the final prediction accuracy. However, little research has been done to investigate how to incorporate additional supervision on the BEV features for better proposal generation. Among the limited number of existing works in this direction, we observe that 3D semantic information has an impressive ability to improve BEV feature qualities. Specifically, several works, including both point-based and grid-based methods~\cite{He_2022_voxset, chen_2022_focal, sun2021rsn, Shi_2019_pointrcnn, chen2022sasa, miao2021pvgnet, xu2021fusionpainting}, propose to integrate foreground-background segmentation predictions into the BEV representation. Instead of the per-point segmentation adopted by these works, Part-$A^2$~\cite{shi_2021_parta2} learns intra-object part locations, which has the additional advantage of utilizing the geometric information from the input point cloud. While beneficial, 3D semantic and geometric supervisions are sparse and fail to compensate for the problem of missing points. Each pixel in the BEV feature map is expected to propose a bounding box center in all commonly used detection heads such as SSD~\cite{Yang_2020_3dssd} and CenterPoint~\cite{Yin_2021_center}. When the 3D input contains very few points around the object center, which is surprisingly common since the LiDAR device only captures a sparse sample of partial surfaces in the environment, the detection head is unable to generate accurate proposals from the BEV features. Figure~\ref{Figure:Intro:dense-sup} presents an illustration of this problem.

Recent research in 3D scene understanding has demonstrated the importance of dense supervision. For example, Yang et al.~\cite{yan2021sparse} improve the single-sweep LiDAR semantic segmentation performance using auxiliary scene completion, while Xu et al.~\cite{xu2021spg} show that generating semantic points that faithfully recover occluded foreground regions enhances object detection accuracy. Unfortunately, obtaining ground-truth annotations of the dense 3D geometry is extremely challenging. To mitigate this issue, we introduce a novel 2D scene representation that encodes the 3D semantics for LiDAR-based object detection. Specifically, consider the direct projection of the 3D scene into the BEV. We model the 2D scene representation as a probability distribution map, where points inside foreground regions are assigned with higher probabilities than the background points. Examples of the proposed semantic scene representation are shown in Figure~\ref{Figure:Teaser}~(a). Unlike conventional segmentation masks, the 2D semantic scene representation consists of real-valued probabilities instead of binary labels. When generated by different methods, the probabilities display different properties. We use different colormaps to show three possible ways to generate the 2D semantic scene representation, including using an explicit network, using an implicit network, and directly projecting ground-truth 3D boxes into 2D. Notably, the 2D semantic scene representation provides dense semantic and geometric information, covering all pixels on the BEV feature map, regardless of the underlying generation method.

A straightforward way to generate the 2D semantic scene representation is to use a discretized image grid. While high-resolution 2D representations are preferable for small objects such as pedestrians, concerns about computational efficiency and memory consumption impede their applications in large 3D scenes. To maintain a careful balance, we combine explicit and implicit representations. We use an explicit network and an implicit network to parallelly generate 2D semantic scenes from the raw BEV features. The refined BEV features are then obtained by fusing the explicit and implicit 2D semantic scenes. We conduct a detailed analysis to reveal how explicit and implicit representations possess complementary properties. Incorporating 2D semantic scene representations require very small modification to existing pipelines. As shown in Figure~\ref{Figure:Teaser}~(b), we introduce an innovative pipeline by adding an extra BEV feature refinement module, which explicitly refines the initial BEV features with 2D dense supervision, referred to as the Semantic Scene Generation Net (SSGNet). 

We integrate SSGNet to various existing LiDAR-based object detectors~\cite{Yin_2021_center, shi2022pillarnet, he2022voxset, Shi_2020_pvrcnn} and evaluate SSGNet on both the Waymo Open Dataset~\cite{Sun_2020_CVPR} and nuScenes~\cite{Caesar_2020_nuscenes}. Extensive experiments demonstrate consistent improvements to all the baseline approaches. SSGNet achieves 78.3\%, 81.1\%, and 74.6\% mAP for the vehicle, pedestrian, and cyclist classes on Waymo (level 1)~\cite{Sun_2020_CVPR}, and 61.7\% mAP for all classes on nuScenes~\cite{Caesar_2020_nuscenes}. Codes will be available at  \href{https://github.com/yanghtr/SSGNet}{https://github.com/yanghtr/SSGNet}.

In summary, we make the following contributions.
\begin{itemize}
\item We introduce a novel 2D semantic scene representation for LiDAR-based 3D object detection.
\item We show how to use explicit and implicit networks to jointly predict the proposed representation.
\item We utilize the predicted 2D semantic scene representation to obtain better BEV features, leading to significant improvement over existing detectors.
\item We validate the benefits of refining BEV features using dense 2D supervision signals and advocate a new detection pipeline with direct BEV supervision.
\end{itemize}


\section{Related Work}
The input to LiDAR-based 3D object detectors is usually represented by point clouds, voxel grids, range images, or their hybrids. We present a survey of recent developments in 3D object detection categorized by the input format.

\noindent\textbf{Point-based methods.}
Point-based methods~\cite{Shi_2019_pointrcnn, Yang_2020_3dssd, Zhang_2022_napae, Qi_2018_frustum, Qi_2019_votenet, Shi_2020_pointgnn, Yang_2019_std} typically apply PointNet and its variants \cite{Qi_2017_pointnet, Qi_2017_pointnet2} on unstructured point clouds to generate proposals from learned point features. Unlike grid-based methods, point-based methods do not suffer from information loss and nicely preserve structural details since they infer from the original LiDAR measurements without any quantization. However, point-based methods use a large number of costly sampling and grouping operations. Compared to methods based on other input representations, point-based methods are also more sensitive to sparsity variations, sampling irregularities, and occlusions.

\noindent\textbf{Grid-based methods.} Grid-based methods~\cite{Zhou_2018_voxelnet, Yan_2018_second, Lang_2019_pointpillar, Deng_2021_voxelrcnn, chen_2022_focal, shi_2021_parta2, Mao_2021_votr} use discrete grids to represent the 3D scene. In addition to straightforward regular voxelization, cylindrical voxelization~\cite{zhou2020cylinder3d} has been proven especially beneficial for semantic segmentation. This paper focuses on regular voxels since they are the most popular choices in object detection. Grid-based methods extract per-voxel features using a sparse convolution backbone~\cite{Yan_2018_second}. To compensate for the information loss due to voxelization, several works~\cite{Lang_2019_pointpillar, Zhou_2018_voxelnet, shi2022pillarnet, He_2022_voxset} additionally utilize a light-weight PointNet~\cite{Qi_2017_pointnet} architecture for voxel feature encoding. Similar to existing works, our method adopts a grid-based representation for the input scene and uses out-of-the-box 3D grid-based backbones to obtain BEV features. We emphasize that existing grid-based methods fail to incorporate an understanding of the scene geometry into the BEV features, which motivates our design.

\noindent\textbf{Range-image-based methods.}
The raw output of a LiDAR sensor is a cylindrical range image. Taking its natural 2D representation as input, range-image-based methods typically apply well-established 2D backbones to extract features before projecting the feature maps to 3D for detection. Several approaches~\cite{meyer2019lasernet,liang2020rangercnn, bewley2020range} leverage the well-understood traditional 2D convolutional network to process range images directly for bounding box parameters regression. Closely related to our work, RSN~\cite{sun2021rsn} uses range images for foreground point segmentation, followed by the projection of the learned features to 3D and the regression of bounding box parameters through 3D sparse convolution. While RSN relies on sparse 3D semantic supervision, our approach uses dense supervision. 

\noindent\textbf{Hybrid representations.}
There is a growing interest in utilizing hybrid input representations for 3D perception tasks. As the pioneer, Frustum PointNets~\cite{qi2018frustum} projects 2D images to the 3D frustum and combines image and point features for object detection. A few other approaches~\cite{Shi_2020_pvrcnn, shi_2021_pvrcnn2, Mao_2021_pyramidrcnn, He_2020_sassd, Noh_2021_hvpr} construct two separate network branches for point and voxel processing and fuse the extracted features for information aggregation. For example, HVPR~\cite{Noh_2021_hvpr} concatenates point-based and voxel-based predictions as the final BEV features. Wang et al.~\cite{wang_2020_pillarod} combine BEV and the cylindrical view to obtain refined pillar features. Their important limitation is the lack of explicit supervision on BEV features. In contrast, we combine the 3D grid representation and 2D semantics, leading to higher efficiency and effectiveness than point-voxel hybrids. 

\section{Approach}

\begin{figure*}[h]
\centering
\includegraphics[width=1.0\textwidth]{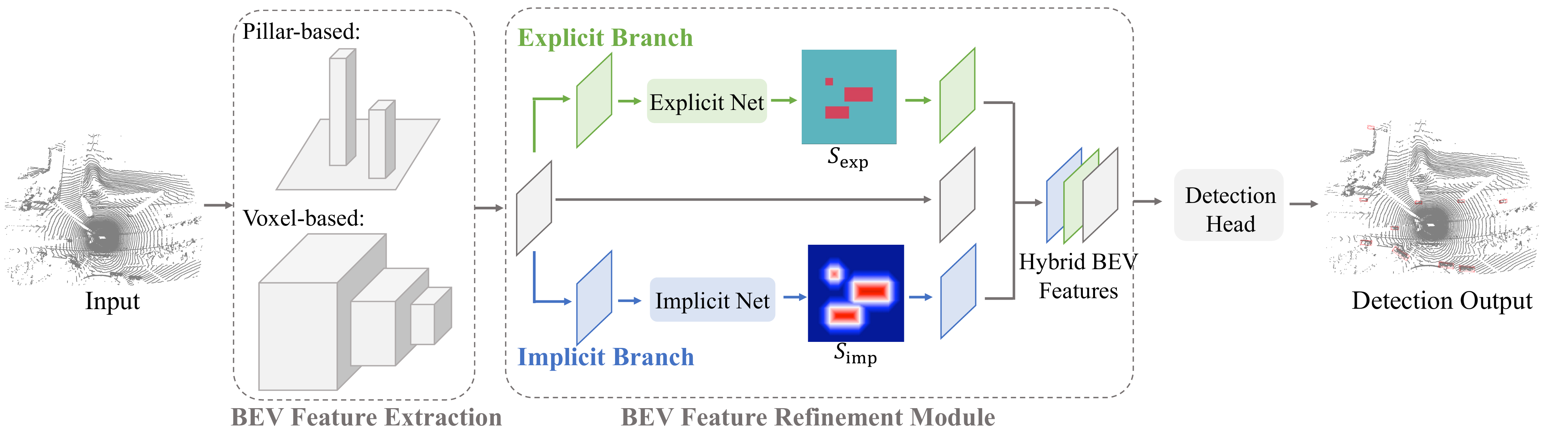}
\caption{SSGNet consists of three modules: the initial BEV feature extraction module, the BEV feature refinement module, and the detection head. The BEV feature refinement module generates hybrid BEV features via 2D semantic scene generation.}
\label{Figure:Pipeline}
\end{figure*}


\subsection{2D Semantic Scene Representation}
\label{Subsec:2D:Semantic:Scene:Rep}
The 2D semantic scene representation is the core of SSGNet, modeled as a 2D probability distribution map:
\[\psi: \mathbb R^2\to\mathbb [0, 1]\]
where each 2D query point $\vq = (x, y)\in \mathbb R^2$ is associated with a probability $p \in [0, 1]$. We assign higher probabilities to points inside foreground regions and lower probabilities to points in the background.

The benefits of the 2D semantic scene representation are three-fold. First, the 2D semantic scene provides an accurate approximation to the \textbf{complete} foreground silhouette in BEV. Contrary to sparse 3D representations such as point clouds and voxels, the 2D semantic scene representation is dense. The dense representation allows a powerful supervision signal covering all locations of interest, including the interior of object cuboids with very few or no LiDAR point captures. In addition, the spatial gradient of the probability distribution is closely related to surface boundaries and edges, which has been shown to enhance object detection thanks to their role in aligning align predicted bounding boxes with geometric cues~\cite{ZhangSYH20}. 

Second, the 2D semantic scene representation naturally models the uncertainty and the structure of the environment. Although conventional segmentation masks are used as training supervision, the proposed 2D scene representation consists of real-valued probabilities instead of binary labels. While regions with high probabilities are more likely associated with objects of interest, a few background locations will have nonzero albeit low probabilities, indicating that the model suspects an object is around even though the confidence is low. Such ambiguities occur when the region of interest (RoI) does not contain a sufficient number of LiDAR points. Without contextual information, it is extremely difficult to accurately detect objects in these regions. By representing the entire scene as one single feature map, the proposed 2D semantic scene representation captures important information about the underlying 3D structure such as the relationship between the RoI and nearby objects.

Third, we point out that training supervision can be easily obtained by the direct projection of 3D bounding boxes into the BEV. This allows the 2D semantic scene representation to connect 3D object detection with 2D image generation, which is a well-studied field with many state-of-the-art design choices.

\subsection{Architecture}
\label{Subsec:Architecture}

As shown in Figure~\ref{Figure:Pipeline}, the proposed SSGNet has a similar pipeline to existing grid-based detectors. Specifically, SSGNet consists of three modules: the initial BEV feature extraction module, the BEV feature refinement module, and the detection head. The main difference to existing pipelines is the use of a novel BEV feature refinement module that incorporates 2D semantic scene generation. 

\noindent\textbf{Initial BEV feature extraction module.}
Given an input scene, the first module in our pipeline extracts initial BEV features using a 3D backbone network. In principle, our SSGNet supports any 3D backbone that regresses BEV features. To demonstrate the generalizability of our design, we experiment with several existing 3D backbones, including CenterPoint~\cite{Yin_2021_center} (sparse 3D Conv), PillarNet~\cite{shi2022pillarnet} (sparse 2D Conv), VoxSeT~\cite{he2022voxset} (transformer), and PV-RCNN~\cite{Shi_2020_pvrcnn} (sparse 3D Conv, two-stage). In these 3D backbone, the input LiDAR points are discretized into small grids before a series of sparse convolution operations or attention layers are applied to extract multi-scale 3D features. The extracted 3D features are then compressed into the 2D BEV features. 

\noindent\textbf{BEV feature refinement module.}
The BEV feature refinement module takes initial BEV features $X\in \mathbb{R}^{h\times w \times d}$ as input, where $h$ and $w$ describe the spatial shape, and $d$ denotes the dimension of the BEV features. The initial BEV features $X$ are simultaneously presented to an explicit network and an implicit network to produce new BEV features $X_\text{exp}$ and $X_\text{imp}$, respectively. These new features are then concatenated with the initial BEV features $X$ to build the enriched hybrid BEV features $F$.

\noindent\textbf{Explicit network.}
The explicit network represents the 2D semantic scene as a rectangular image grid $S_\text{exp}\in [0, 1]^{h\times w \times 1}$, whose spatial resolution is the same as the initial BEV features $X$. The explicit network formulates 2D semantic scene generation as a classical 2D image regression problem. While SSGNet supports any out-of-the-box image generation network, our experiments apply the popular U-Net architecture~\cite{Ronneberger_2015_MICCAI} for simplicity. Consider:
\begin{align}
S_\text{exp} & = f_\text{exp}(X), \label{Eq:exp:gen-s} \\
X_\text{exp} & = X \oplus g_\text{exp} (S_\text{exp}), \label{Eq:exp:gen-x}
\end{align}
where $f_\text{exp}$ denotes the U-Net, $g_\text{exp}$ is a Conv-BN-ReLU layer that lifts the feature dimension from $1$ to $d$, and $\oplus$ stands for channel-wise concatenation.

By interpreting the network output as real-valued probabilities, the 2D semantic scene representation differs from a binary segmentation mask. Keeping the probabilities allows the enriched BEV features to preserve visual information in low-confidence regions. While a perfect binary segmentation is certainly desirable, predicting such a segmentation from the initial BEV features is practically impossible due to the problem of missing points. As an alternative, we use continuous probabilities instead of discrete binary labels to model the 2D semantic scene. Experimentally, Section~\ref{Section:Ablation} demonstrates the effectiveness of this design choice through an ablation study.

We use 2D ground-truth segmentation masks to supervise all pixels densely. These ground-truth segmentation masks can be easily obtained by projecting the ground-truth bounding boxes from 3D into 2D. We adopt the focal loss~\cite{Lin_2022_CVPR} to calculate the difference between the predicted and ground-truth segmentations since the foreground pixels usually take up only a tiny fraction of the $h \times w$ grid.

\begin{figure}
\centering
\includegraphics[width=0.4\textwidth]{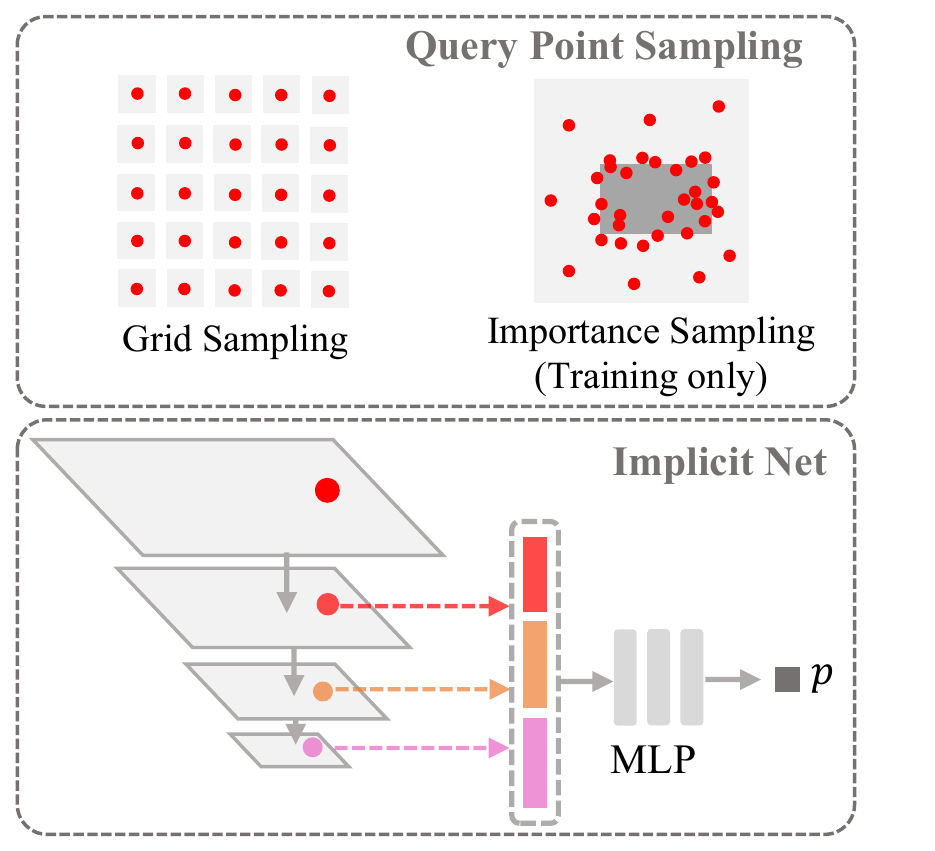}
\caption{While implicit network training uses both grid and importance samplings, only grid sampling is used in testing.}
\label{Figure:ImpNet}
\end{figure}

\noindent\textbf{Implicit network.} Inspired by the recent advances in neural implicit representations~\cite{Park:CVPR:2019, Chen_2019_CVPR, Mescheder_2019_CVPR}, we propose to learn another instance of the probability distribution map using an implicit network. The architecture of the implicit network is shown in Figure~\ref{Figure:ImpNet}. In addition to 2D coordinates, the implicit network $\phi$ takes a latent code as input and uses the latent code as the source of scene-level information. Specifically, given the initial BEV features $X$, we first embed $X$ into a latent space:
\begin{align}
    L = h_\text{imp}(X),
\end{align}
where $h_\text{imp}$ is a Conv-BN-ReLU layer, and $L$ is the latent code. Notably, $L$ is not required to have the same spatial shape as $X$. Our experiments demonstrate that increasing the spatial resolution of $L$ contributes to higher detection accuracy albeit introducing an overhead on GPU memory usage. The 2D semantic scene is represented implicitly as:
\begin{align}
    p = \phi(\vq, L),
\end{align}
where $\vq\in \mathbb{R}^2$ is the query point in BEV and $p\in [0, 1]$ is the probability of $\vq$ being a foreground location.

To obtain both local and global information, we extract multi-scale features from $L$. For each 2D query point $\vq$, we locate its position in the multi-scale feature maps and extract per-point features through bilinear interpolation, which are then concatenated and fed to an MLP network to predict the probability $p$.

An appealing property of the implicit 2D semantic scene representation is that importance sampling can be naturally combined in the training process. The use of importance sampling has been repeatedly proven effective in LiDAR-based 3D object detection~\cite{Zhang_2022_napae, chen_2022_focal}. Intuitively, we expect the model to sample more points around the foreground objects because they provide good guidance to object discovery.


\begin{table*}
    \centering
    \footnotesize
    \caption{Quantitative comparison of SSGNet with different baselines on the Waymo Open Dataset~\cite{Sun_2020_CVPR} (val split).}
    \label{tab:exp:results:waymo_val}
    \begin{Tabular}{c|c|cc|cc|cc|cc|cc|cc} 
    \hline
    \multicolumn{14}{c}{Results When Trained With 20\% Training Examples} \\
    \hline
    \multirow{2}{*}{Methods} & \multirow{2}{*}{Stages} & \multicolumn{2}{c|}{Veh. (L1)}  & \multicolumn{2}{c|}{Veh. (L2)}  & \multicolumn{2}{c|}{Ped. (L1)} & \multicolumn{2}{c|}{Ped. (L2)} & \multicolumn{2}{c}{Cyc. (L1)} & \multicolumn{2}{c}{Cyc. (L2)}    \\
              &     & mAP & mAPH & mAP & mAPH & mAP & mAPH & mAP & mAPH & mAP & mAPH & mAP & mAPH \\
    \hline
    SECOND~\cite{Yan_2018_second}  & One & 71.0 & 70.3 & 62.6 & 62.0 & 65.2 & 54.2 & 57.2 & 47.5 & 57.1 & 55.6 & 55.0 & 53.5 \\
    PointPillars~\cite{Lang_2019_pointpillar} & One & 70.4 & 69.8 & 62.2 & 61.6 & 66.2 & 46.3 & 58.2 & 40.6 & 55.3 & 51.8 & 53.2 & 49.8 \\
    IA-SSD~\cite{Zhang_2022_napae} & One & 70.5 & 69.7 & 61.6 & 60.8 & 69.4 & 58.5 & 60.3 & 50.7 & 67.7 & 65.3 & 65.0 & 62.7 \\
    SST (Center)~\cite{fan_2022_sst} & One & 75.1 & 74.6 & 66.6 & 66.2 & 80.1 & 72.1 & 72.4 & 65.0 & 71.5 & 70.2 & 68.9 & 67.6 \\
    \hline
    CenterPoint-Voxel~\cite{Yin_2021_center} & One & 72.8 & 72.2 & 64.9 & 64.4 & 74.2 & 68.0 & 66.0 & 60.3 & 71.0 & 69.8 & 68.5 & 67.3 \\ 
    \cite{Yin_2021_center}+SSGNet (Ours) & One & 75.0 & 74.4 & 67.0 & 66.5 & 75.6 & 69.0 & 67.9 & 61.9 & 71.8 & 70.7 & 69.2 & 68.1 \\
    \hline
    \multicolumn{2}{c|}{\textbf{Absolute Improvements}} & \textbf{+2.2} & \textbf{+2.2} & \textbf{+2.1} & \textbf{+2.1} & \textbf{+1.4} & \textbf{+1.0} & \textbf{+1.9} & \textbf{+1.6} & \textbf{+0.8} & \textbf{+0.9} & \textbf{+0.7} & \textbf{+0.8} \\
    \hline
    PillarNet~\cite{shi2022pillarnet} & One & 74.0 & 73.4 & 66.0 & 65.5 & 72.3 & 63.8 & 64.5 & 56.7 & 67.8 & 66.5 & 65.2 & 64.0 \\ 
    \cite{shi2022pillarnet}+SSGNet (Ours) & One & 75.8 & 75.1 & 67.9 & 67.3 & 74.0 & 65.3 & 66.1 & 58.2 & 69.3 & 68.1 & 66.7 & 65.5 \\ 
    \hline
    \multicolumn{2}{c|}{\textbf{Absolute Improvements}} & \textbf{+1.8} & \textbf{+1.7 }& \textbf{+1.9} & \textbf{+1.8} & \textbf{+1.7} & \textbf{+1.5} & \textbf{+1.6} & \textbf{+1.5} & \textbf{+1.5} & \textbf{+1.6} & \textbf{+1.5} & \textbf{+1.5} \\
    \hline
    VoxSeT~\cite{he2022voxset} & One & 72.1 & 71.6 & 63.6 & 63.2 & 77.9 & 69.6 & 70.2 & 62.5 & 69.9 & 68.5 & 67.3 & 66.0 \\
    \cite{he2022voxset}+SSGNet (Ours) & One & 76.0 & 75.5 & 67.9 & 67.4 & 79.4 & 71.3 & 71.8 & 64.3 & 72.1 & 70.9 & 69.4 & 68.3 \\ 
    \hline
    \multicolumn{2}{c|}{\textbf{Absolute Improvements}} & \textbf{+3.9} & \textbf{+3.9} & \textbf{+4.3} & \textbf{+4.2} & \textbf{+1.5} & \textbf{+1.7} & \textbf{+1.6} & \textbf{+1.8} & \textbf{+2.2} & \textbf{+2.4} & \textbf{+2.1} & \textbf{+2.3} \\
    \hline
    PV-RCNN (Center)~\cite{Shi_2020_pvrcnn} & Two & 76.0 & 75.4 & 68.0 & 67.5 & 75.9 & 69.4 & 67.7 & 61.6 & 70.2 & 69.0 & 67.7 & 66.6 \\
    \cite{Shi_2020_pvrcnn}+SSGNet (Ours) & Two & 78.3 & 77.8 & 70.1 & 69.6 & 79.6 & 73.3 & 71.4 & 65.4 & 71.5 & 70.4 & 68.9 & 67.9 \\
    \hline
    \multicolumn{2}{c|}{\textbf{Absolute Improvements}} & \textbf{+2.3} & \textbf{+2.4} & \textbf{+2.1} & \textbf{+2.1} & \textbf{+3.7} & \textbf{+3.9} & \textbf{+3.7} & \textbf{+3.8} & \textbf{+1.3} & \textbf{+1.4} & \textbf{+1.2} & \textbf{+1.3} \\
    \hline
    \hline
    \multicolumn{14}{c}{Results When Trained With 100\% Training Examples} \\
    \hline
    \multirow{2}{*}{Methods} & \multirow{2}{*}{Stages} & \multicolumn{2}{c|}{Veh. (L1)}  & \multicolumn{2}{c|}{Veh. (L2)}  & \multicolumn{2}{c|}{Ped. (L1)} & \multicolumn{2}{c|}{Ped. (L2)} & \multicolumn{2}{c|}{Cyc. (L1)} & \multicolumn{2}{c}{Cyc. (L2)}    \\
              &     & mAP & mAPH & mAP & mAPH & mAP & mAPH & mAP & mAPH & mAP & mAPH & mAP & mAPH \\
    \hline
    AFDetV2~\cite{hu2022afdetv2} & One & 77.6 & 77.1 & 69.7 & 69.2 & 80.2 & 74.6 & 72.2 & 67.0 & 73.7 & 72.7 & 71.0 & 70.1 \\
    SWFormer~\cite{sun2022swformer} & One & 77.8 & 77.3 & 69.2 & 68.8 & 80.9 & 72.7 & 72.5 & 64.9 & - & - & - & - \\
    CenterFormer~\cite{zhou2022centerformer} & One & 75.0 & 74.4 & 69.9 & 69.4 & 78.6 & 73.0 & 73.6 & 68.3 & 72.3 & 71.3 & 69.8 & 68.8 \\
    PV-RCNN~\cite{Shi_2020_pvrcnn} & Two & 78.0 & 77.5 & 69.4 & 69.0 & 79.2 & 73.0 & 70.4 & 64.7 & 71.5 & 70.3 & 69.0 & 67.8 \\
    Part-A2-Net~\cite{shi2020points} & Two & 77.1 & 76.5 & 68.5 & 68.0 & 75.2 & 66.9 & 66.2 & 58.6 & 68.6 & 67.4 & 66.1 & 64.9 \\
    LiDAR-RCNN~\cite{li2021lidar} & Two & 76.0 & 75.5 & 68.3 & 67.9 & 71.2 & 58.7 & 63.1 & 51.7 & 68.6 & 66.9 & 66.1 & 64.4 \\
    \hline
    VoxSeT~\cite{he2022voxset} & One & 74.5 & 74.0 & 66.0 & 65.6 & 80.0 & 72.4 & 72.5 & 65.4 & 71.6 & 70.3 & 69.0 & 67.7 \\
    \cite{he2022voxset}+SSGNet (Ours) & One & 78.3 & 77.8 & 70.2 & 69.8 & 81.1 & 74.1 & 73.6 & 67.1 & 74.6 & 73.4 & 71.9 & 70.7 \\
    \hline
    \multicolumn{2}{c|}{\textbf{Absolute Improvements}} & \textbf{+3.8} & \textbf{+3.8} & \textbf{+4.2} & \textbf{+4.2} & \textbf{+1.1} & \textbf{+1.7} & \textbf{+1.1} & \textbf{+1.7} & \textbf{+3.0} & \textbf{+3.1} & \textbf{+2.9} & \textbf{+3.0} \\
    \hline
    \end{Tabular}
\end{table*}

\begin{table*}
    \centering
    \footnotesize
    \caption{Quantitative comparison of SSGNet with different baselines on the nuScenes dataset~\cite{Caesar_2020_nuscenes}.}
    \label{tab:exp:results:nuscenes}
    \begin{Tabular}{c|cc|c|c|c|c|c|c|c|c|c|c}
    \hline
    Methods & mAP & NDS & Car & Truck & Bus & Trailer & C.V. & Ped & Mot & Byc & T.C. & Bar \\
    \hline
    SECOND-CBGS~\cite{zhu_2019_cbgs} & 50.6 & 62.3 & - & - & - & - & - & - & - & - & - & - \\
    PillarNet-18~\cite{shi2022pillarnet} & 59.9 & 67.4 & - & - & - & - & - & - & - & - & - & -  \\
    LargeKernel3D~\cite{chen_2022_largeKernels} & 60.5 & 67.6 & 85.8 & 59.0 & 72.8 & 40.0 & 19.2 & 85.6 & 61.3 & 43.6 & 70.4 & 67.5 \\
    Focals Conv~\cite{chen_2022_focal} & 61.2 & 68.1 & 86.6 & 60.2 & 72.3 & 40.8 & \textbf{20.1} & \textbf{86.2} & 61.3 & 45.6 & 70.2 & \textbf{69.3} \\
    CenterPoint-Voxel~\cite{Yin_2021_center} & 59.0 & 66.4 & 85.6 & 57.2 & 71.2 & 37.3 & 16.2 & 85.1 & 58.4 & 41.0 & 69.2 & 68.2 \\
    \hline
    \cite{Yin_2021_center}+SSGNet (Ours) & \textbf{61.7} & \textbf{68.3} & \textbf{87.1} & \textbf{60.5} & \textbf{73.9} & \textbf{43.3} & 19.2 & \textbf{86.2} & \textbf{63.0} & \textbf{45.7} & \textbf{72.1} & 66.4 \\
    \hline
    \end{Tabular}
\end{table*}

Computing the implicit training loss requires us to sample a large number of query points and compare the ground-truth mask and predicted probabilities, which is a time-consuming operation and typically involves additional data pre-processing~\cite{Park:CVPR:2019}. We introduce a handy yet efficient point sampler built upon hyper-parameters $N_s\in \mathbb{N}$, $\alpha\in [0, 1]$, and $\beta\in [0, 1]$. Here, $N_s$ is the total number of points to be sampled, and $\alpha$ controls the ratio of uniform sampling to importance sampling. After obtaining $\alpha N_s$ points by regular uniform sampling across the entire scene, the other $(1 - \alpha ) N_s$ points, referred to as \textit{important points}, are sampled either inside or in close proximity to the foreground object bounding boxes. The number of points inside foreground boxes and the number of points outside but close to foreground boxes are balanced by the hyper-parameter $\beta$. Let $K\in \mathbb{N}$ be the number of foreground objects in a specific scene. We first uniformly sample $\frac{(1-\alpha)N_s}{K}$ points in a unit cube for each bounding box. The $k$-th box ($k=\{1, 2, ..., K\}$) can be parameterized with $(c_x^{(k)}, c_y^{(k)}, c_z^{(k)}, s_x^{(k)}, s_y^{(k)}, s_z^{(k)}, r^{(k)})$, with elements representing the center, size, and orientation. We slightly enlarge the size $\sqrt{\frac{1}{\beta}}$ times and generate a new box with parameters $(c_x^{(k)}, c_y^{(k)}, c_z^{(k)}, \sqrt{\frac{1}{\beta}}s_x^{(k)}, \sqrt{\frac{1}{\beta}}s_y^{(k)}, s_z^{(k)}, r^{(k)})$. We then scale the previously sampled points in the unit cube according to these new box parameters. An important property of this sampling method is that all foreground objects contain the same number of point samples, regardless of their respective sizes. Among all the important points, there are at least $(1-\alpha)\beta$ samples inside the bounding boxes. Our experiments set $\alpha=\frac{1}{3}$ and $\beta=\frac{2}{3}$. Uniform sampling (controlled by $\alpha$) in the whole scene also produces a few points inside the bounding boxes. The percentage of foreground and background point samples are both close to 50\%, leading to a nice balance between contextual and conceptual information. Notably, we point out that the sampling process discussed in this paragraph only occurs during the training phase as an intermediate step for loss computation. 

The detection head expects the BEV features to be represented on a rectangular grid and not by an implicit function. During both the training and testing stages, we query all locations with integer coordinates on a $h\times w$ grid to obtain an explicit 2D probability distribution map $S_\text{imp}\in \mathbb{R}^{h\times w}$. Similar to the explicit representation, we generate the implicit BEV features $X_\text{imp}$ through a Conv-BN-ReLU layer $g_\text{imp}$ and channel-wise concatenation:
\begin{align}
    X_\text{imp} = X \oplus g_\text{imp} (S_\text{imp}).
\end{align}

\noindent\textbf{Detection head.}
After obtaining the refined BEV features, we can use any existing detection head to generate object proposals. While our experiments focus on a few existing models~\cite{Yin_2021_center, shi2022pillarnet, he2022voxset, Shi_2020_pvrcnn} due to their state-of-the-art performance, we point out that SSGNet is also compatible with other designs.
\section{Experiments}

\begin{table*}
\footnotesize
\begin{center}
\caption{Ablation study on the Waymo Open Dataset~\cite{Sun_2020_CVPR} (val split).}
\label{tab:exp:results:ablation}
\begin{Tabular}{c|cc|cc|cc|cc|cc|cc} 
\hline
\multirow{2}{*}{Method}  & \multicolumn{2}{c|}{Veh. (L1)}  & \multicolumn{2}{c|}{Veh. (L2)}  & \multicolumn{2}{c|}{Ped. (L1)} & \multicolumn{2}{c|}{Ped. (L2)} & \multicolumn{2}{c|}{Cyc. (L1)} & \multicolumn{2}{c}{Cyc. (L2)}  \\ 
                     & mAP & mAPH & mAP & mAPH & mAP & mAPH & mAP & mAPH & mAP & mAPH & mAP & mAPH \\
\hline
CenterPoint-Voxel           & 72.8 & 72.2 & 64.9 & 64.4 & 74.2 & 68.0 & 66.0 & 60.3 & 71.0 & 69.8 & 68.5 & 67.3 \\
CenterPoint-Voxel$\times 2$ & 72.9 & 72.3 & 65.0 & 64.5 & 74.0 & 67.8 & 66.0 & 60.3 & 71.2 & 70.0 & 68.6 & 67.4 \\ 
CenterPoint-Voxel$\times 3$ & 73.1 & 72.6 & 65.0 & 64.5 & 73.9 & 67.7 & 66.0 & 60.4 & 71.3 & 70.2 & 68.7 & 67.6 \\ 
\hline
SSGNet-Voxel-Explicit           & 74.5 & 74.0 & 66.5 & 66.0 & 75.4 & 69.0 & 67.6 & 61.8 & 71.4 & 70.2 & 69.0 & 67.9 \\
SSGNet-Voxel-Implicit           & 74.1 & 73.6 & 66.0 & 65.5 & 74.8 & 68.3 & 66.9 & 61.0 & 71.1 & 69.9 & 68.7 & 67.6 \\ 
SSGNet-Voxel-Hybrid-Binary      & 74.1 & 73.5 & 66.0 & 65.5 & 75.1 & 68.6 & 67.3 & 61.3 & 70.2 & 69.0 & 67.9 & 66.7 \\ 
\hline
SSGNet-Voxel                    & \textbf{75.0} & \textbf{74.4} & \textbf{67.0} & \textbf{66.5} & \textbf{75.6} & \textbf{69.0} & \textbf{67.9} & \textbf{61.9} & \textbf{71.8} & \textbf{70.7} & \textbf{69.2} & \textbf{68.1} \\ 
\hline
\end{Tabular}
\end{center}
\end{table*}

\subsection{Experimental Setup}
\label{Section:Dataset}

Waymo Open Dataset~\cite{Sun_2020_CVPR} is a large-scale dataset. We use version 1.2, which contains 158,081 training samples and 39,987 validation samples. There are two difficulties: levels 1 and 2, where the bounding boxes contain at least five and one LiDAR points, respectively. We use official metrics, including the mean Average Precision (mAP) and the mean Average Precision weighted by Heading (mAPH). We benchmark multi-class 3D object detection using only single-frame LiDAR point clouds. The IoU thresholds for vehicles, pedestrians, and cyclists are 0.7, 0.5, and 0.5.

nuScenes~\cite{Caesar_2020_nuscenes} is another popular benchmark dataset with a total of 1,000 driving sequences. Compared to Waymo~\cite{Sun_2020_CVPR}, nuScenes includes more object categories, covering different types of vehicles such as cars, trucks, buses, and trailers. We use the mean Average Precision (mAP) and the nuScenes Detection Score (NDS) to compare different baseline approaches.

\subsection{Implementation Details}
\label{Section:Implementation}

We integrate SSGNet into four different top-performing object detectors, including CenterPoint~\cite{Yin_2021_center}, PillarNet~\cite{shi2022pillarnet}, VoxSeT~\cite{he2022voxset}, and PV-RCNN~\cite{Shi_2020_pvrcnn}. We use a detection range of $\textup{[-75.2m, 75.2m]}$ for the X and Y axes and $\textup{[-2m, 4m]}$ for the Z axis. The voxel sizes are $\textup{(0.1m, 0.1m, 0.15m)}$. All of our implementations are built upon the OpenPCDet~\cite{openpcdet2020} code base. Following Zhang et al.~\cite{Zhang_2022_napae}, we mainly use 20\% training frames but evaluate on whole validation set due to limited computation resources. This setup applies to all baselines. We provide more comprehensive evaluation using 100\% training examples against VoxSeT~\cite{he2022voxset}.

For Waymo Open Dataset, we train the networks on 8 Tesla V100 GPUs for 30 epochs using the ADAM optimizer~\cite{KingmaB14} with the one-cycle policy. For nuScenes, we train the networks for 20 epochs. All methods use the same data pre-processing, augmentation, and post-processing steps. To establish fair comparisons, we keep all other settings as the default in OpenPCDet with class-agnostic NMS enabled.

\subsection{Analysis of Results}
\label{Section:Results}
Table~\ref{tab:exp:results:waymo_val} presents a summary of the quantitative evaluation on the Waymo Open Dataset~\cite{Sun_2020_CVPR}. We choose four recent top-performing object detectors, including three single-stage methods (CenterPoint~\cite{Yin_2021_center}, PillarNet~\cite{shi2022pillarnet}, VoxSeT~\cite{he2022voxset}) and one two-stage method (PV-RCNN). We add a 2D semantic scene generation module to each of these detectors and compare the prediction quality with the primitive models. Additionally, we present the results from the state-of-the-art point-based detector that does not involve BEV features, IA-SSD~\cite{Zhang_2022_napae}, and three other baseline methods~\cite{Yan_2018_second, Lang_2019_pointpillar, Zhang_2022_napae}.

\begin{figure}
\centering
\begin{overpic}[width=0.45\textwidth]{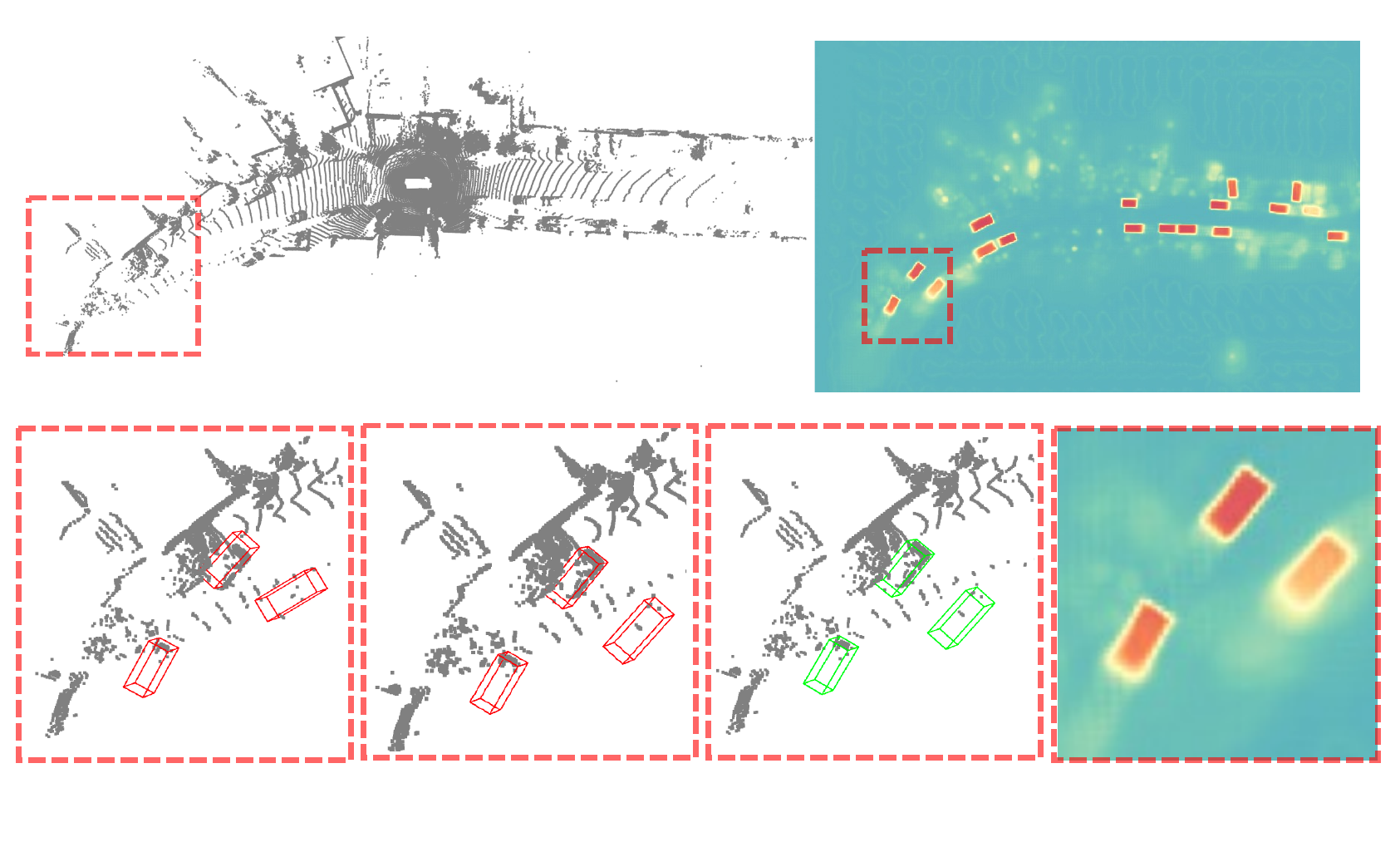}
\put(1,2) {CenterPoint~\cite{Yin_2021_center}}
\put(35,2){Ours}
\put(60,2){GT}
\put(82,2){2DSS}
\end{overpic}

\caption{A failure cause in CenterPoint~\cite{Yin_2021_center} successfully corrected by SSGNet. Contextual information provided by 2D semantic scene (2DSS) helps to infer the bounding box parameters from sparse input regions.}
\label{Figure:Exp:analysis2}
\end{figure}
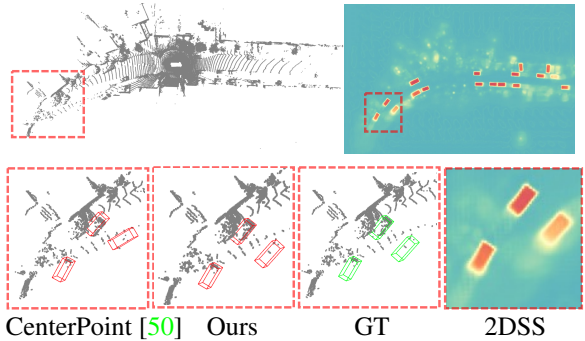

The integration of SSGNet demonstrates consistent improvements from all baseline methods. Under the mAP metric, our method improves CenterPoint~\cite{Yin_2021_center} (the voxel-based sparse convolutional variant) by 2.2\%, 1.4\%, and 0.8\% for level-1 vehicles, pedestrians, and cyclists, respectively. Compared to PillarNet~\cite{shi2022pillarnet}, the absolute improvements are 1.8\%, 1.7\%, and 1.5\%. For VoxSeT~\cite{he2022voxset}, the improvements are 3.9\%, 1.5\%, and 2.2\% using 20\% training data, and 3.8\%, 1.1\%, and 3.0\% using 100\% training data. Compared to PV-RCNN~\cite{Shi_2020_pvrcnn}, the incorporation of SSGNet leads to improvements of 2.3\%, 3.7\%, and 1.3\%. Overall, we observe a more significant improvement in the vehicle class than the pedestrian and cyclist classes in the one-stage detectors. Our explanation is that the larger size of a typical vehicle results in a longer average distance from its center to the surface, making vehicles more susceptible to the problem of missing points. As shown in Figure~\ref{Figure:Exp:analysis2}, even in regions with very sparely sampled points, the proposed 2D semantic scene representation provides a faithful description of the object silhouettes thanks to the density of our supervision. Additionally, Figure~\ref{Figure:Exp:analysis2} suggests that contextual information plays an important role. While our detected boxes follow a similar orientation, the detection from CenterPoint lacks such uniformity, leading to inaccuracies and lower IoUs. 

In addition to Waymo ~\cite{Sun_2020_CVPR}, Table~\ref{tab:exp:results:nuscenes} reports the quantitative results on nuScenes~\cite{Caesar_2020_nuscenes}. Integrating SSGNet into CenterPoint~\cite{Yin_2021_center} achieves an mAP of 61.7\% and an NDS of 68.3\%, representing 2.7\% and 1.9\% improvements from the primitive model. This outperforms the state-of-the-art Focals Conv~\cite{chen_2022_focal} introduced in CVPR 2022.

\begin{figure*}
\centering
\begin{overpic}[width=0.8\textwidth]{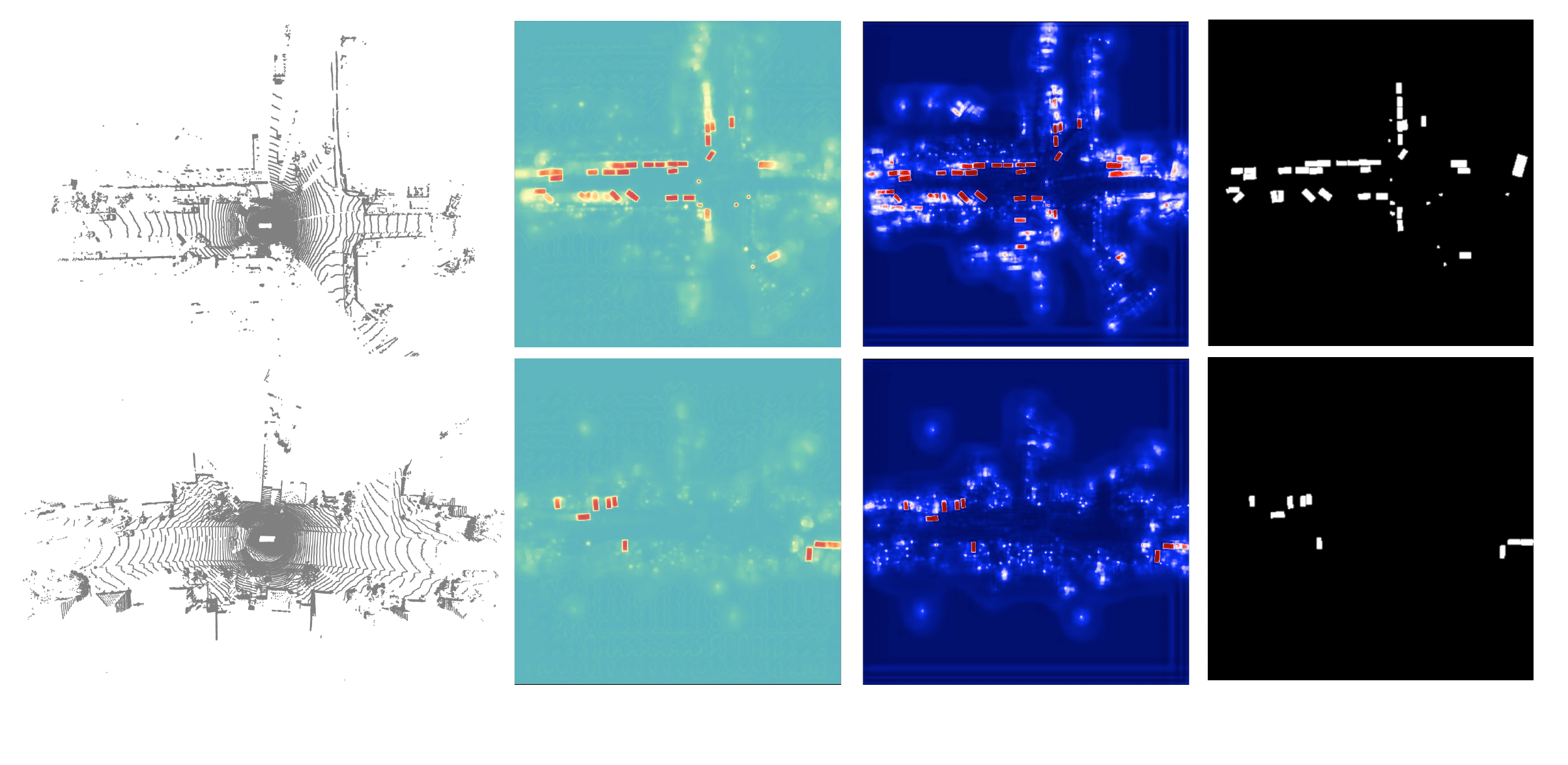}
\put(15,2) {Input}
\put(37,2){Explicit 2DSS}
\put(60,2){Implicit 2DSS}
\put(83,2){GT 2DSS}
\end{overpic}

\caption{Additional visualizations of 2D semantic scenes (2DSS) generated via different methods.}
\label{Figure:Exp:2DSS-vis}
\end{figure*}

\subsection{Ablation Study}
\label{Section:Ablation}

\noindent\textbf{Effect of incorporating 2D semantic scene generation.} 
Starting from a basic CenterPoint~\cite{Yin_2021_center} model, we gradually introduce 2D semantic scene generation networks. In Table~~\ref{tab:exp:results:ablation}, SSGNet-Voxel-Explicit refers to the CenterPoint model after adding an explicit network, and SSGNet-Voxel-Implicit denotes the CenterPoint model after adding an implicit network. To compensate for the introduction of more trainable parameters, we compare the modified models with CenterPoint-Voxel$\times2$, whose network size has been proportionally increased. Specifically, CenterPoint-Voxel$\times2$ uses two spare backbones to generate two sets of features and concatenates them as the final BEV feature.

We observe that using either of the two modified networks consistently improves the primitive model across all three object classes, although the explicit network shows a slightly greater boost. As suggested by Figure~\ref{Figure:Exp:2DSS-vis}, the implicit network occasionally assigns high probabilities to background regions, creating false positives.

\noindent\textbf{Effect of hybrid representation.} 
In Table~\ref{tab:exp:results:ablation}, the comparisons of SSGNet-Voxel-Explicit and SSGNet-Voxel-Implicit with the hybrid model (SSGNet-Voxel) demonstrate that the implicit and explicit representations have complementary benefits to accurate object detection.
As shown in Figure~\ref{Figure:Exp:2DSS-vis}, if we generate object proposals based on the predicted probabilities, the sensitivity of the implicit network will lead to a high recall, and the robustness of the explicit network will contribute to a high precision. In addition, the implicit scene contains valuable background information, such as the boundaries associated with land and walls. 

\noindent\textbf{Effects of probability distribution modeling.}
By interpreting the network output as real-valued probabilities, the 2D semantic scene representation differs from a binary segmentation mask. While a perfect binary segmentation is certainly desirable, predicting such a segmentation from the initial BEV features is practically impossible due to the problem of missing points. As an alternative, we use continuous probabilities instead of discrete binary labels to model the 2D semantic scene. As shown in Table~\ref{tab:exp:results:ablation}, SSGNet-Voxel-Hybrid-Binary denotes a variant of the full model (SSGNet-Voxel) where the 2D semantic scene representations are converted to binary masks under a threshold of 0.5. We observe that this binarization leads to a noticeable performance drop. This suggests that the probabilistic modeling naturally balances the number of object proposals with uncertainties. Leaving the ambiguities for the detection head to resolve is empirically better than using an arbitrary threshold to eliminate the ambiguities, which leads to an inevitable information loss. 

\noindent\textbf{Effects of supervision density.}
The primitive VoxSeT~\cite{he2022voxset} has already utilized sparse per-point semantic segmentation. We modify VoxSeT+SSGNet by disabling the per-point segmentation and obtain close performance. Due to the space constraint, we refer interested readers to the supplementary material for more details about how it demonstrates the importance of dense supervision.

\section{Conclusions and Future Work}
This paper introduces the Semantic Scene Generation Net (SSGNet), which encodes dense semantic and geometric information into Bird's-Eye View (BEV) features under a hybrid of explicit and implicit representations. Experiments and detailed analysis demonstrate that SSGNet can be easily integrated into existing LiDAR-based 3D object detectors and achieve significant improvements in accuracy. In particular, the density of semantic scene supervision is an important contributor to the performance gain. In the future, we plan to investigate other possible dense supervision signals. Another promising direction is to explore whether there is a better 2D representation for the 3D semantics. 

\noindent\textbf{Acknowledgement.} We would like to acknowledge NSF IIS-2047677, HDR-1934932, and CCF-2019844.

{\small
\bibliographystyle{ieee_fullname}
\bibliography{egbib}
}

\newpage
\clearpage
\appendix
The supplementary materials provide more results on the effects of supervision density in Section \ref{sec:supp:results:density}, and additional implementation details in Section \ref{sec:supp:network}.

\section{Effects of Supervision Density}
\label{sec:supp:results:density}

We provide an ablation study on the dense supervision in Table~\ref{tab:supp:results:ablation}. VoxSeT~\cite{he2022voxset} has already utilized per-point semantic segmentation, which relies on sparse supervision. After directly adding the proposed BEV feature refinement module with dense supervision, VoxSeT+SSGNet improves VoxSeT significantly ($+3.9\%$ Veh. L1, $+4.3\%$ Veh. L2). We then remove the per-point segmentation head from VoxSeT+SSGNet and obtain a new model VoxSeT-NoSparseSeg+SSGNet. From the results in Table~\ref{tab:supp:results:ablation}, VoxSeT-NoSparseSeg+SSGNet has very close performance to VoxSeT+SSGNet, which indicates that the dense supervision, instead of the sparse supervision, is the key to the performance gains.

\begin{table*} 
\footnotesize
\begin{center}
\caption{Ablation study on the Waymo Open Dataset (val split, $20\%$ data).}
\label{tab:supp:results:ablation}
\begin{Tabular}{c|cc|cc|cc|cc|cc|cc} 
\hline
\multirow{2}{*}{Method}  & \multicolumn{2}{c|}{Veh. (L1)}  & \multicolumn{2}{c|}{Veh. (L2)}  & \multicolumn{2}{c|}{Ped. (L1)} & \multicolumn{2}{c|}{Ped. (L2)} & \multicolumn{2}{c|}{Cyc. (L1)} & \multicolumn{2}{c}{Cyc. (L2)}  \\ 
                     & mAP & mAPH & mAP & mAPH & mAP & mAPH & mAP & mAPH & mAP & mAPH & mAP & mAPH \\
\hline
VoxSeT~\cite{he2022voxset} & 72.1 & 71.6 & 63.6 & 63.2 & 77.9 & 69.6 & 70.2 & 62.5 & 69.9 & 68.5 & 67.3 & 66.0 \\
\cite{he2022voxset}+SSGNet  & 76.0 & 75.5 & 67.9 & 67.4 & 79.4 & 71.3 & 71.8 & 64.3 & 72.1 & 70.9 & 69.4 & 68.3 \\ 
\cite{he2022voxset}-NoSparseSeg+SSGNet& 76.1 & 75.6 & 67.9 & 67.5 & 79.2 & 71.4 & 71.8 & 64.5 & 72.4 & 71.2 & 69.7 & 68.5 \\ 
\hline
\end{Tabular}
\end{center}
\end{table*}

\section{Implementation Details}
\label{sec:supp:network}

\subsection{More Details of the Implicit Network}
 The query points from importance sampling is only used to compute loss during training and will not be used by the detection head. During training and testing, the query points from grid sampling will be fed to the implicit network to generate a grid of probability values $S_\text{imp}$. Specifically, the loss of implicit network is
\vspace{-0.05in}
\begin{equation}
   L_\text{imp} = \frac{1}{|Q_\text{IS}|} \sum_{\bs{q} \in Q_\text{IS}} L_\text{focal}(\phi(\bs{q}, L), p_\text{gt}),
\end{equation}
\vspace{-0.02in}
where $Q_\text{IS}$ is the set of points from importance sampling, $L_\text{focal}(\cdot, \cdot)$ is the focal loss, $p_\text{gt}$ is the ground truth probability, $\phi$ is the implicit network. In addition,
\begin{equation}
   S_\text{imp} = \left[p_{ij}\right]_{h\times w}, p_{ij} = \phi(\bs{q}_{ij}, L), \bs{q}_{ij} \in Q_\text{GS},
\end{equation}
where $Q_\text{GS}$ is the set of points from grid sampling. Note that there are no differences in generating implicit feature $S_\text{imp}$ during training and testing (both from grid sampling). 

\subsection{Overall Network Details}
In this section, we provide more details about the designs of CenterPoint-Voxel + SSGNet in the Waymo dataset. Similar designs also apply to other architectures. For the two-stage model PV-RCNN, we only modify its first stage. The voxel size is (0.1m, 0.1m, 0.15m) and the input spatial grid shape is [1504, 1504] for the X and Y axes. We use the down-sampled features with XY spatial grid shape [376, 376] as input to the BEV feature refinement module. 

In the explicit network (U-Net), the input features are down-sampled four times. In the implicit network, the input features are down-sampled three times , and 3 MLP layers are used in the decoder. Note that although both the  encoders of the explicit and implicit network use multi-scale features, the decoders are distinct (ConvNet \textit{vs.} MLP), which contributes to the complementary properties of the generated explicit and implicit 2D semantic scene.

The overall loss is
\begin{align}
    L_\text{total} = L_\text{rpn} + L_\text{exp} + \lambda_\text{imp} L_\text{imp},
\end{align}
where $L_\text{rpn}$ is the overall loss used in CenterPoint-Voxel, $L_\text{exp}$ and $L_\text{imp}$ are the losses for the explicit and implicit 2D semantic scene generation, respectively. We set $\lambda_\text{imp} = 5$.

\end{document}